\documentclass[twoside,11pt]{article}

\usepackage{jmlr2e}
\usepackage{booktabs}
\usepackage{bm}
\usepackage{MyMathCmds}

\usepackage{MarkBiblatexCmds}  
\usepackage{cleveref}
\usepackage{todonotes}

\usepackage{pifont}
\newcommand{\cmark}{\ding{51}}%
\newcommand{\xmark}{\ding{55}}%

\jmlrheading{1}{2021}{}{}{}{Dutordoir et al.$\quad$\textsuperscript{$*$}  Work done while at Secondmind.}

\ShortHeadings{GPflux: A Library for Deep Gaussian Processes}{Dutordoir et al.}
\firstpageno{1}

\newcommand{\mat}[1]{\mathbf{{#1}}} 
\renewcommand{\vec}[1]{\bm{{#1}}}
\newcommand{\MS}{\mat{S}}
\newcommand{\MK}{\mat{K}}
\newcommand{\Eye}{\mat{I}}
\newcommand{\MSigma}{\mat{\Sigma}}

\newcommand{\Kuu}{\MK_{\vu_{\ell} \vu_{\ell}}}

\newcommand{\ku}{\vec{k}_{\vu_{\ell}}}

\usepackage{float}
\usepackage{listings}

\newfloat{lstfloat}{htbp}{lop}
\floatname{lstfloat}{Listing}

\usepackage{color}

\definecolor{codegreen}{rgb}{1,0.69,0}
\definecolor{codegray}{rgb}{0.5,0.5,0.5}
\definecolor{codepurple}{rgb}{0.58,0,0.82}
\definecolor{backcolour}{rgb}{0.97,0.97,0.97}
 
\lstdefinestyle{mycodestyle}{
    basicstyle=\ttfamily\scriptsize,
    backgroundcolor=\color{backcolour},   
    commentstyle=\color{codegreen},
    numberstyle=\tiny\color{codegray},
    breakatwhitespace=false,         
    breaklines=true,                 
    captionpos=b,                    
    numbers=left,                    
    numbersep=5pt,                  
    showspaces=false,                
    showstringspaces=false,
    showtabs=false,                  
    tabsize=1
}

\begin{document}
\newcommand{\inlinecode}{\texttt}

\title{GPflux: A Library for Deep Gaussian Processes}

\author{\name Vincent Dutordoir\email vd309@cam.ac.uk \\
       \addr University of Cambridge, Cambridge, UK \\
                Secondmind, Cambridge, UK
       \AND
       \name Hugh Salimbeni \email hugh.salimbeni@gmail.com  \\
       \addr Secondmind, Cambridge, UK
       \AND
       \name Eric Hambro \email eric.hambro@gmail.com \\
       \addr Secondmind, Cambridge, UK
       \AND
       \name John McLeod \email johnangusmcleod@gmail.com \\
       \addr Secondmind, Cambridge, UK
       \AND
       \name Felix Leibfried \email felix.leibfried@gmail.com \\
       \addr Secondmind, Cambridge, UK
       \AND
       \name Artem Artemev \email a.artemev20@imperial.ac.uk \\
       \addr Imperial College London, London, UK \\ 
                Secondmind, Cambridge, UK
       \AND
       \name Mark van der Wilk\textsuperscript{$*$} \email m.vdwilk@imperial.ac.uk \\
       \addr Imperial College London, London, UK
       \AND
       \name James Hensman \email james.hensman@gmail.com \\
       \addr Secondmind, Cambridge, UK
       \AND
       \name Marc P. Deisenroth\textsuperscript{$*$} \email m.deisenroth@ucl.ac.uk \\
       \addr University College London, London, UK
       \AND
       \name ST John \email st@secondmind.ai \\
       \addr Secondmind, Cambridge, UK
}

\editor{}

\maketitle

\begin{abstract}%
We introduce GPflux, a Python library for Bayesian deep learning with a strong emphasis on deep Gaussian processes (DGPs). Implementing DGPs is a challenging endeavour due to the various mathematical subtleties that arise when dealing with multivariate Gaussian distributions and the complex bookkeeping of indices. To date, there are no actively maintained, open-sourced and extendable libraries available that support research activities in this area. GPflux aims to fill this gap by providing a library with state-of-the-art DGP algorithms, as well as building blocks for implementing novel Bayesian and GP-based hierarchical models and inference schemes. GPflux is compatible with and built on top of the Keras deep learning eco-system. This enables practitioners to leverage tools from the deep learning community for building and training customised Bayesian models, and create hierarchical models that consist of Bayesian and standard neural network layers in a single coherent framework. GPflux relies on GPflow for most of its GP objects and operations, which makes it an efficient, modular and extensible library, while having a lean codebase.
\end{abstract}

\begin{keywords}
Bayesian deep learning, deep Gaussian processes, TensorFlow and GPflow
\end{keywords}

\section{Introduction}

Deep neural networks (DNNs) are flexible parametric function approximators that can be used for supervised and unsupervised learning, especially in applications where there is an abundance of data, such as in computer vision \citep{alexnet}, natural language processing \citep{attention}, and planning \citep{Schrittwieser2020MasteringAG}. However, in small-data and noisy settings, DNNs can overfit \citep{Goodfellow}, or simply make overconfident predictions, which prevents their use in safety-critical applications \citep{yuan2019adversarial}. To address this, Bayesian learning algorithms propose to replace the usual point estimates of parameters with probability distributions that quantify the uncertainty that remains due to a lack of data. The resulting Bayesian neural networks (BNNs) can be more robust to overfitting, and make predictions together with a measure of their reliability. 

Alternatively, instead of representing probability distributions in weight space, Gaussian processes (GPs) can be used to represent uncertainty directly in function space \citep{rasmussen2006}. \citet{Damianou2013} first used Gaussian processes as layers to create the Deep Gaussian Process (DGP), which enables learning of more powerful representations through compositional features. Recent work has shown that DGPs are particularly promising because function-space Bayesian approximations seem to be of higher quality than those of weight-space BNNs \citep{foong2019expressiveness,damianou2015thesis}.

\section{Motivation}
Despite their advantages, DGPs have not been adopted nor studied as widely as (Bayesian) DNNs. A possible explanation for this is the non-existence of well-designed and extendable software libraries that underpin research activities. This is crucial, especially for DGPs, as implementing them is a challenging endeavour, even when relying on toolboxes that support automatic differentiation. This is due to, among other things, the implicit (and infinite) basis functions in GPs, the difficulty of keeping track of indices in the multi-layered, multi-output setting and the numerous numerical implementation subtleties when conditioning Gaussian distributions.

To date, there are no actively maintained, open-sourced and extendable DGP libraries available. Some packages (see \Cref{tab:existing}) exist, but they are written with a single use-case in mind, and only implement one variation of a model or inference scheme. None of them address the fact that much of the code in DGPs can be factored out and reused for building novel DGP models --- these are exactly the abstractions GPflux is providing. Building on top of GPflow, GPflux is also designed to be efficient and modular. That is, the library allows new variants of models and approximations to be implemented without modifying the core GPflux source \citep{gpflow,van2020framework}.

The aim of GPflux is twofold. First, it aims to provide researchers with reusable components to develop new DGP models and inference schemes. Second, it aims to provide practitioners with existing state-of-the-art (deep) GP algorithms and layers, such as latent variables, convolutional and multi-output models. GPflux is built on top of TensorFlow\citep{tensorflow,tensorflowprobability} and is compatible with Keras \citep{keras}. This makes it possible to leverage on a plethora of tools developed by the deep learning community for building, training and deploying deep learning models.

\section{Deep Gaussian Processes: Brief Overview of Model and Inference}
Given a dataset $\{(\vx_i, y_i)\}_{i=1}^N$, a Deep Gaussian process \citep[DGP]{Damianou2013} is built by composing several GPs, where the output of one layer is fed as input to the next. For each layer $f_\ell(\cdot)$, we assume that it is a-priori distributed according to a GP with a kernel $k_\ell(\cdot, \cdot)$. The DGP is then defined as the composition $\mathcal{F}(\cdot) = f_L(\ldots f_2(f_1(\cdot)))$. We refer to the latent function evaluation of a datapoint $\vx_i$ at the $\ell^{\text{th}}$ GP as $\vh_{i, \ell} = f_\ell(\vh_{i, \ell-1})$ with $\vh_{i, 0} = \vx_i$.
We further assume a general likelihood $y_i \given \mathcal{F}, \vx_i \sim p(y_i \given \mathcal{F}(\vx_i))$.

\citet{salimbeni2017doubly} introduced an elegant and scalable inference scheme for DGPs based on the work of Sparse Variational GPs \citep{hensman2015scalable}. It defines $L$ independent approximate posterior GPs $q(f_\ell(\cdot))$ of the form 
\begin{equation}
\label{eq:qf}
    q(f_\ell(\cdot)) = \GP\Big( \underbrace{\ku(\cdot) \Kuu^{-1} \vm_{\ell}}_{=: \vmu_\ell(\cdot)};
\quad 
\underbrace{k_{\ell}(\cdot, \cdot) + \ku^\top(\cdot)\Kuu^{-1}(\MS_{\ell} - \Kuu) \Kuu^{-1} \ku(\cdot)}_{=: \MSigma_\ell(\cdot)}\Big),
\end{equation}
where $\{\vm_\ell, \MS_\ell\}_{\ell=1}^L$ parameterises the variational approximate posterior $q(\vu_{\ell})$ over inducing variables. $\Kuu$ and $\ku(\cdot)$ are covariance matrices computed using the kernel $k_\ell(\cdot, \cdot)$. The model is trained by optimising a lower bound (ELBO) on the log marginal likelihood
\begin{equation}
\label{eq:elbo}
\log p\left(\{y_i\}_{i=1}^N\right) \ge \underbrace{\sum\nolimits_i \Exp{q(\vh_{i, L})}{\log p(y_i \given h_{i, L})}}_{\text{data-fit}}
 - \underbrace{\sum\nolimits_{\ell} \KL{q(\vu_\ell)}{p(\vu_\ell)}}_{\text{complexity}}.
\end{equation}
The complexity term in the ELBO can be computed in closed form because all of the distributions are Gaussian. An end-to-end differentiable and unbiased Monte-Carlo estimate of the data-fit term can be computed with samples from $q(\vh_{i, L})$, which can be obtained by iteratively propagating datapoints through the layers using the reparametrisation trick: $\vh_{\ell} = \vmu_{\ell}(\vh_{\ell - 1}) + \sqrt{\MSigma(\vh_{\ell - 1})}\,\bm{\epsilon}$ with $\bm{\epsilon} \sim \NormDist{0, \Eye}$. We refer the interested reader to \citet{van2020framework} and \citet{Leibfried2020Tutorial} for in-depth discussion of this method.

\section{Key Features and Design}


GPflux is designed as a deep learning library where functionality is packed into layers, and layers can be stacked on top of each other to form a hierarchical (i.e.\ deep) model. Next, we focus on the layers and subsequently show how to create models using these layers. We briefly highlight some useful tooling provided by Keras for training these Bayesian models.

\subsection{Layers}

The key building block in GPflux is the \texttt{GPlayer}, which represents the prior and posterior of a single (multi-output) GP, $f_\ell(\cdot)$. It can be seen as the analogue of a standard fully-connected (dense) layer in a DNN, but with an infinite number of basis functions. It is defined by a \texttt{Kernel}, \texttt{InducingVariables}, and \texttt{MeanFunction}, which are all GPflow objects. Adhering to the Keras design, a layer has to implement a \texttt{call} method which usually maps a Tensor to another Tensor. A \texttt{GPLayer}'s \texttt{call} is slightly different in that it takes the output of the previous layer, say $\vh_{\ell-1}$, but returns an object that represents the complete Gaussian \emph{distribution} of $f_{\ell}(\vh_{\ell-1})$ as given by \cref{eq:qf}. If a subsequent layer is not able to use the previous layer's distributional output, a sample will be taken using the reparametrisation trick. This functionality is provided by TensorFlow Probability's \texttt{DistributionLambda} layer \citep{tensorflowprobability}.

GPflux also provides other Bayesian and GP-based layers. A \texttt{LatentVariableLayer} implements a layer which augments its inputs $\vh_{i, \ell-1}$ with latent variables  $\bm{w}_i$, usually through concatenation. This leads to more flexible DGPs that can model complex, non-Gaussian densities \citep{Dutordoir2018,Salimbeni2019}. \texttt{Convolutional} layers can be used for temporal or spatially structured data \citep{vanderwilk2017conv,Dutordoir2020convolutional}. GPflux also provides non-GP-specific layers, such as \texttt{BayesianDenseLayer} which implements a dense layer for variational Bayesian neural networks. Moreover, thanks to the compatibility of GPflux with the Keras eco-system, it is possible to naturally combine GPflux layers with standard DNN components, such as convolutional or fully-connected layers, as shown in Listing~1. This variety of different building blocks provided by GPflux in a single unified framework paves the way for systematic evaluation of these methods.

\begin{lstfloat}[t]
\begin{lstlisting}[language=python, style=mycodestyle]
# Initialise a 4-layer model consisting of NN layers and GP layers
model = Sequential([Dense(...), Convolution(...), GPLayer(...), GPLayer(...)])
model.compile(loss=LikelihoodLoss(Gaussian()), optimizer="Adam")
# Fitting
callbacks = [ReduceLROnPlateau(), TensorBoard(), ModelCheckpoint()]
model.fit(X, Y, callbacks=callbacks)
# Evaluating
model.predict(X)
\end{lstlisting}
\caption{Initialising, fitting and evaluating a GPflux hybrid NN-DGP model.}
\end{lstfloat}

\subsection{Models and Fitting}

As shown in the second line of Listing~1, we can in most cases directly make use of Keras' \texttt{Sequential} to combine the different GPflux and Keras layers into a hierarchical model. This can be convenient because we limit the number of wrappers around our core layer functionality. However, certain GPflux layers (e.g.,~\texttt{LatentVariableLayer}) require both features $\{\vx_i\}$ \emph{and} the target $\{y_i\}$ in training, which is a functionality that Keras does not provide directly. For these use-cases GPflux provides the specialised \texttt{DeepGP} class.

Deep neural networks (DNNs) are trained by minimising the prediction discrepancy for examples in a training set. This is a similar to the data-fit term in the ELBO (\cref{eq:elbo}), which is passed to the framework using \texttt{LikelihoodLoss(Gaussian())} in the listing. The KL complexity terms of the ELBO are added to the loss by the \texttt{GPLayer} calls.

GPflux enables DGP models to reuse much of the tooling developed by the deep learning community. E.g., during training it can be advantageous to use Keras' \texttt{ReduceLROnPlateau} to lower the learning rate when the ELBO converges. Other callbacks make it possible to monitor the optimisation trace using TensorBoard or save the optimal weights to disk --- many of these features have not been leveraged in (deep) GP libraries before. Finally, adhering to the Keras interface also gives GPflux models a battle-tested interface (e.g.,~\texttt{fit}, \texttt{predict}, \texttt{evaluate}) which should ease its adoption in downstream applications.

\section{Final Remarks}
GPflux is a toolbox dedicated to Bayesian deep learning and Deep Gaussian processes. GPflux uses the mathematical building blocks from GPflow and combines these with the powerful layered deep learning API provided by Keras. This combination leads to a framework that can be used for: (i) researching new models, and (ii) building, training and deploying Bayesian models in a modern way.

A number of steps have been taken to ensure the quality and usability of the project. All GPflux source code is available at \url{http://github.com/secondmind-labs/GPflux/}. We use continuous integration and have a test code coverage of over 97\%. To learn more or get involved we encourage the reader to have a look at our documentation which contains tutorials and a thorough API reference.

\acks{We want to thank Nicolas Durrande, Carl Rasmussen, Dongho Kim, and everybody else at Secondmind who was involved in the open-sourcing effort.}

\appendix
\section{Existing open-source efforts}

\begin{table}[H]
    \centering
    \resizebox{\textwidth}{!}{\begin{tabular}{lllc}
\toprule
Package &
Implements & 
Last Commit &
Code Tests
\\ \midrule
SheffieldML/PyDeepGP &
\cite{Damianou2013,dai2015variational} &
Nov 2018 &
\xmark 
\\
FelixOpolka/Deep-Gaussian-Process &
\cite{Salimbeni2019} &
Mar 2021 &
\xmark
\\
ICL-SML/Doubly-Stochastic-DGP &
\cite{salimbeni2017doubly,salimbeni2018natural} &
Feb 2019 &
\cmark
\\
hughsalimbeni/DGPs\_with\_IWVI &
\cite{Salimbeni2019} &
May 2019 &
\cmark
\\
cambridge-mlg/sghmc\_dgp &
\cite{havasi2018inference} &
Feb 2019 &
\xmark
\\
kekeblom/DeepCGP &
\cite{blomqvist2019deep} &
Sep 2019 &
\cmark
\\
GPyTorch/DeepGP module &
\cite[adapted to Conj. Gr.]{salimbeni2017doubly} &
Jul 2020 &
\cmark
\\
\bottomrule
\end{tabular}}
    \caption{A summary of existing Python Deep GP libraries at the time of writing. \label{tab:existing}}
\end{table}

\setlength\bibitemsep{1.5\itemsep}
\printbibliography


\end{document}